\documentclass[sigconf]{acmart}

\acmDOI{10.1145}

\usepackage{natbib}
\usepackage[utf8]{inputenc}
\usepackage{dblfloatfix}
\usepackage{algorithmic}
\usepackage{graphicx}
\usepackage{textcomp}
\usepackage{xcolor}
\def\BibTeX{{\rm B\kern-.05em{\sc i\kern-.025em b}\kern-.08em
    T\kern-.1667em\lower.7ex\hbox{E}\kern-.125emX}}
\begin{document}

\title{Expert-Driven Monitoring of Operational ML Models}

\begin{abstract}
We propose \textit{Expert Monitoring}, an approach that leverages domain expertise to enhance the detection and mitigation of concept drift in machine learning (ML) models. Our approach supports practitioners by consolidating domain expertise related to concept drift-inducing events, making this expertise accessible to on-call personnel, and enabling automatic adaptability with expert oversight.
%-- contributing to a new avenue of research: \textit{human-centered model monitoring}.
\end{abstract}

\author{Joran Leest}
\affiliation{%
  \institution{\textit{Vrije Universiteit Amsterdam}}
  \country{The Netherlands}
  \orcid{0009-0003-1335-6559}
}
\email{j.g.leest@vu.nl}

\author{Claudia Raibulet}
\affiliation{%
  \institution{\textit{Vrije Universiteit Amsterdam}}
  \country{The Netherlands}
  \orcid{0000-0002-7194-3159}
}
\email{c.raibulet@vu.nl}

\author{Ilias Gerostathopoulos}
\affiliation{%
  \institution{\textit{Vrije Universiteit Amsterdam}}
  \country{The Netherlands}
  \orcid{0000-0001-9333-7101}
}
\email{i.g.gerostathopoulos@vu.nl}

\author{Patricia Lago}
\affiliation{%
  \institution{\textit{Vrije Universiteit Amsterdam}}
  \country{The Netherlands}
  \orcid{0000-0002-2234-0845}
}
\email{p.lago@vu.nl}

\settopmatter{authorsperrow=4}

%\title{Expert Monitoring: Human-Centered Concept Drift Detection in Machine Learning Operations}

%\title{Expert-Driven Monitoring of Operational ML Models}

\maketitle

\section{Introduction}
The ubiquitous integration of machine learning (ML) in modern software systems marks a shift from deterministic behavior of software to behavior derived from stochastic processes. This transition brings a significant challenge: ensuring the consistent performance of ML models that are subject to changes in the data distribution, due to external events and data integrity issues \cite{huyen2022}. This phenomenon, known as \textit{data drift}, encompasses \textit{feature drift} (changes in the input distribution, $P(X)$) and \textit{concept drift} (changes in the conditional probability distribution of the target variable given an input, $P(Y|X)$)~\cite{webb2016characterizing}. Notably,
concept drift has a substantial influence on model performance
and necessitates mitigative action \cite{cloudera2021, gama2014survey}.

The growing use of ML underscores the importance of addressing concept drift, particularly in sensitive areas like credit card fraud detection, to prevent discriminatory actions \cite{pombal2022understanding}. We must scrutinize concept drift mitigation solutions, including insights from data mining \cite{gama2014survey, lu2018}, and integrate them into the emerging ML operations (MLOps) framework used by practitioners \cite{kreuzberger2023machine}.

In this paper, we explore MLOps practices and challenges, highlighting the inherent limitations of automated concept drift detection and mitigation, and motivate the the necessity of domain expertise (Section 2). Based on this understanding, we outline an approach to integrate expert knowledge into a monitoring system (Section 3). Finally, we discuss the key aspects of our approach (Section 4), and conclude by outlining our future plans (Section 5).
%We conclude by reflecting on our approach and the novel research direction of human-centered model monitoring to which it contributes. 
%Moreover, we outline future plans for further development and validating our approach (see Section V).

\section{Navigating the Darkness: Concept Drift in Practice}
This section discusses the challenge of detecting and mitigating concept drift, and how domain expertise is utilized to address it.
% where the reliance on automated techniques by practitioners remains sparse\cite{shankar2022operationalizing,shergadwala2022human}
%This exploration is informed by recent seminal works on MLOps.

\subsection{The Practical Challenges of Concept Drift Detection -- Where the Shadows Lie} % which challenges do practioners experience
We see the challenges in detecting and deciphering concept drift's latent aspects, often leaving practitioners "wandering in the dark".
% more descriptive (we see the challenges)

\subsubsection{Concept Drift Detection Without Labeled Data}
Unlike feature drift, which can be readily observed in input data and its effects on model performance inferred through various estimation methods \cite{rabanser2019failing, sethi2017reliable, shankar2021towards, chen2021mandoline, garg2022leveraging, guillory2021predicting}, concept drift detection depends on monitoring metrics such as accuracy \cite{gama2014survey, huyen2022}. This is challenging in real-world scenarios, as labeled data is often delayed or completely absent \cite{huyen2022}. For example, in e-commerce churn prediction, churn determination only occurs after a specified time frame (e.g., a month or year) \cite{ahn2020survey}.

In the absence of labeled data, practitioners typically rely on detecting drift in the model's predictions and features, such as through the use of a Kolmogorov-Smirnov test, which can act as a proxy to infer the presence of concept drift and its effect on model performance \cite{cloudera2021, breck2017ml, huyen2022}. This strategy makes sense because it leverages the common co-occurrence of feature drift and concept drift \cite{cloudera2021,gama2014survey, liu2023handling}. In real-world scenarios, events impact how data is generated, collected, and managed, leading to changes in our models that capture these processes \cite{groh2022identifying}. Feature drift can act as a visible signal of these changes. % INSTEAD??: In practice, events (e.g. national holidays) can change how data is generated, collected, and managed, leading to distribution shifts in variables underlying these processes\cite{groh2022identifying}. Consequently, shifts in variables used as model features can lead to feature drift, while shifts in latent variables can result in concept drift. Despite the latent nature of concept drift, feature drift can serve as a visible signal of its occurrence. 
Nonetheless, inferring concept drift from feature drift remains challenging, as their presence or severity does not always correlate \cite{cloudera2021}. As a result, triggering alerts for every instance of feature drift generates many false alarms \cite{shankar2022operationalizing, shankar2021towards, shergadwala2022human}.

To illustrate the challenge of concept drift detection without labeled data, let us consider a model that predicts whether a customer will churn based on the customer age and recent page visits (Fig. 1).

\vspace{-0.8em}

\begin{center}
    \begin{figure}[!htpb]
    \centering
    \label{fig:drift_visual}
    \includegraphics[width=0.478\textwidth]{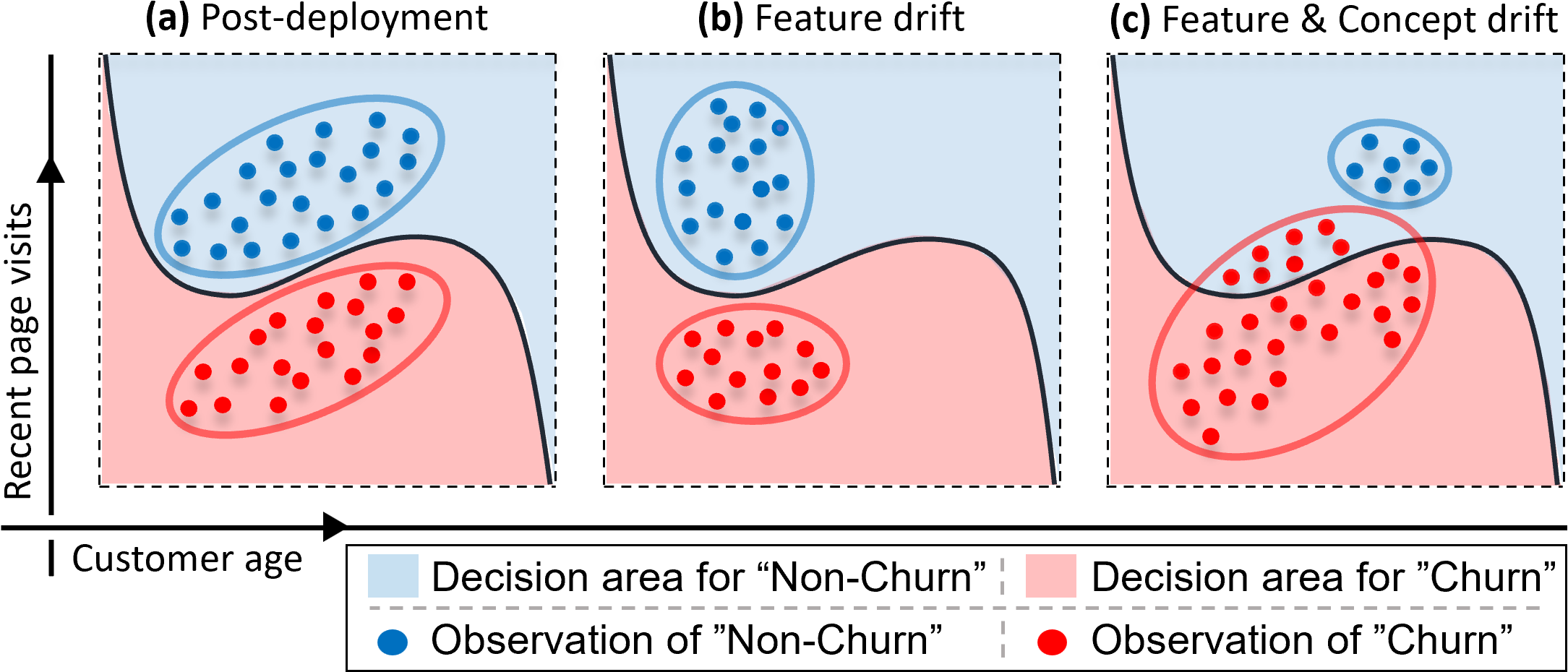}
    \caption{Data drift in a customer churn prediction model.}
    \end{figure}
\end{center}

\vspace{-1.5em}

In Fig. 1, (a) depicts the post-deployment observations and the learned decision boundary (represented by the black line). After deployment, two events occur: (b) the web shop launches a marketing campaign targeting young people, causing drift in the customer age feature, and (c) a competitor's marketing campaign for a new product line aimed at young people makes the web shop's young customers switch to the competitor's service, causing drift in the recent page visits feature. Event (b) does not affect customer satisfaction; the learned decision boundary remains valid. Conversely, event (c) negatively affects the (latent) customer satisfaction among the web shop's younger customers, whom have become aware of the new offering. Consequently, concept drift occurs, rendering the learned mapping function invalid. The core problem emerges: \textbf{in the absence of labeled data, monitoring systems that rely on feature drift detection do not discern event (b) from (c)}.

\subsubsection{Deciphering the Nature of Concept drift Post-Detection}
Even when the presence of concept drift can be confidently inferred, effective response selection requires an understanding of the detected drift's characteristics \cite{gama2014survey,han2022survey}. This includes the drift severity, recurrence, duration, and transition speed \cite{webb2016characterizing}. For example, in the case of recurrent drift, reactivating a previous model version might effectively resolve it \cite{mallick2022matchmaker}, whereas abrupt drift might necessitate a complete model retrain. Conversely, for a short-lived drift duration, retraining the model is not desirable; instead, it might suffice to temporarily fall back on a more simple model.

Comprehending the nature of concept drift after its detection (e.g. whether it is a recurring event) remains a significant challenge \cite{shergadwala2022human}, with current concept drift detection methods falling short in facilitating comprehension of the drift's characteristics \cite{lu2018}.

\subsection{Domain Expertise -- A Light in Dark Places, When All Other Lights Go Out} % what do practioners do
In recent works, Shankar et al. \cite{shankar2022operationalizing} and Shergadwala et al. \cite{shergadwala2022human} conducted insightful interview studies involving ML engineers. These studies highlighted the practical aspects of monitoring ML models and the strategies employed by ML engineers in practice. Their findings showed a common theme: \textbf{automated concept drift detection and mitigation is not a predominant tool in the arsenal of practitioners. Instead, human intervention and on-call rotations play a crucial role in monitoring ML models}. In doing so, practitioners are actively involved in maintaining the model's overall and subgroup performances, aiming to optimize business value and ensure the model's fairness, respectively \cite{ shankar2022operationalizing, huyen2022}.

Now, let us first revisit the challenges we pinpointed in the previous section and consider how the domain expertise of a human-in-the-loop can address them. Afterwards, we examine the difficulties that arise when relying on domain expertise for model monitoring.

\subsubsection{Addressing Concept Drift with Domain Expertise}
To demonstrate how domain expertise can aid in detecting and mitigating concept drift, we again consider customer churn prediction and provide an illustrative example (see Fig. 2). Here, we present three illustrative instances of multivariate feature drift, along with an expert's assessment of the potential underlying event and the presence of concept drift. Subsequently, we describe an appropriate response tailored to the nature of the detected drift.

\begin{center}
    \begin{figure}[!htpb]
    \centering
    \label{fig:events}
    \includegraphics[width=0.478\textwidth]{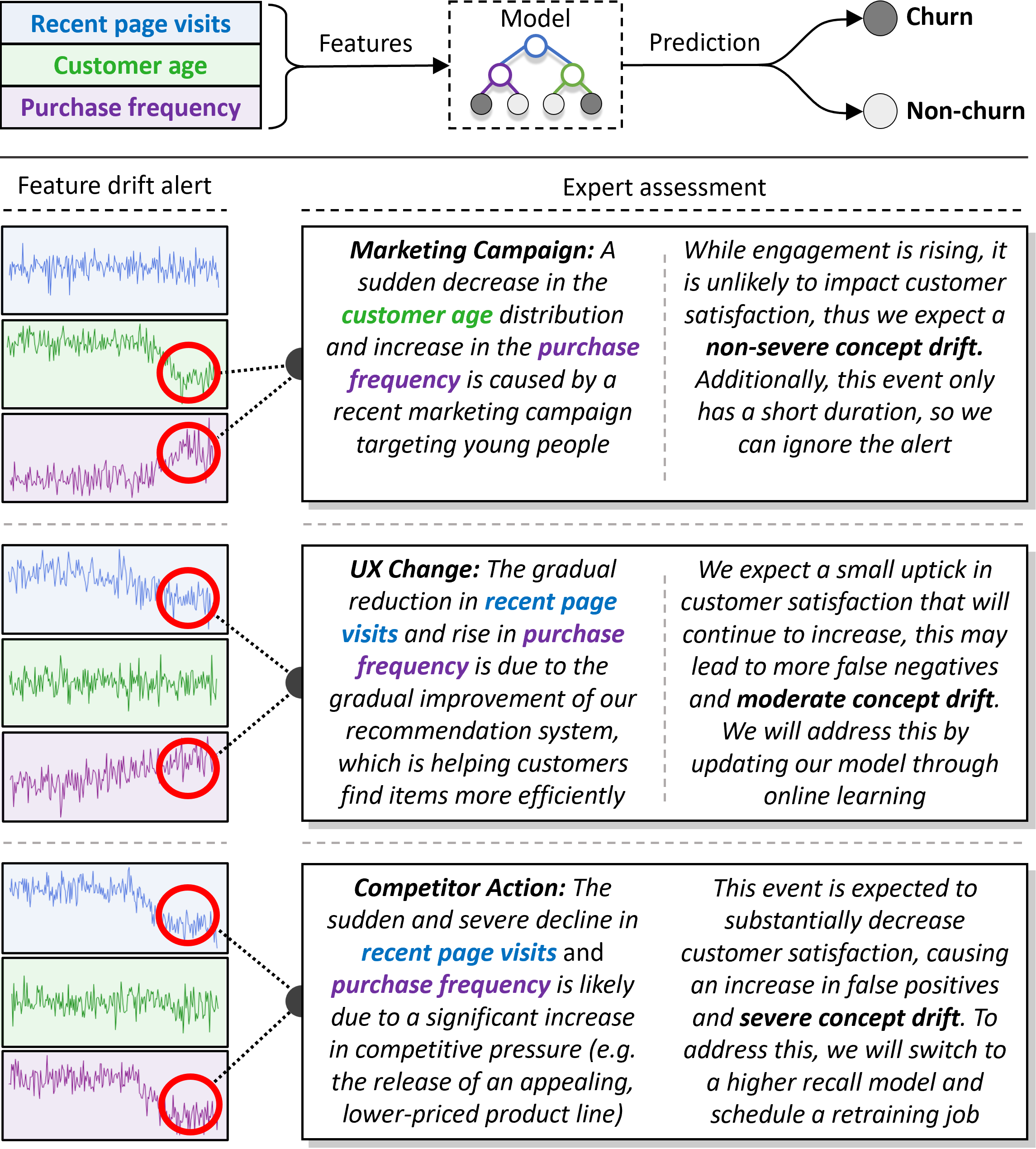}
    \caption{An illustrative case for customer churn prediction, showing expert assessments for three cases of feature drift. }
    \end{figure}
\end{center}

\vspace{-2em}

The events depicted in Fig. 2 emphasize the vital importance of domain expertise in the process of concept drift detection and mitigation. While all events were effectively identified through feature drift detection, they varied significantly in their impact on model performance and the proper course of action.

\subsubsection{Why Is It Hard to Rely on Domain Expertise?}
In the example (Fig. 2), we assumed the assessments were conducted by an expert with a deep understanding of the model and its application context. In practice, this responsibility often falls to on-call ML engineers, who oversee the monitoring of ML models \cite{shankar2022operationalizing, shergadwala2022human}. These models are often developed by different teams and operate in various contexts -- processes that ML engineers have little to no involvement in! This proves indeed challenging, as quoted by an ML engineer: \textit{"The pain point is dealing with that alert fatigue and the domain expertise necessary to know when to take action when on-call"} \cite{shankar2022operationalizing}. 

In addition to the need for domain expertise, ML engineers have also expressed a need for centralized model governance \cite{shergadwala2022human}, as knowledge about ML models and their application context is often dispersed and inaccessible \cite{neutatz2021cleaning}. This issue becomes particularly difficult to manage for organizations that run numerous models, each with its unique feature set and context. For example, in addition to churn prediction, organizations might use models for tasks like demand prediction, product recommendation, and personalized search. The issue of acquiring and retaining domain expertise are exacerbated by factors such as staff turnover, limited documentation, and the need for extensive training \cite{shergadwala2022human, shankar2022operationalizing}. These challenges highlight the need to consolidate domain expertise and make it accessible to on-call ML engineers.

\vspace{-2em}

%\vspace{-1em}
\begin{center}
    \begin{figure*}[!b]
    \centering
    \includegraphics[width=0.999\textwidth]{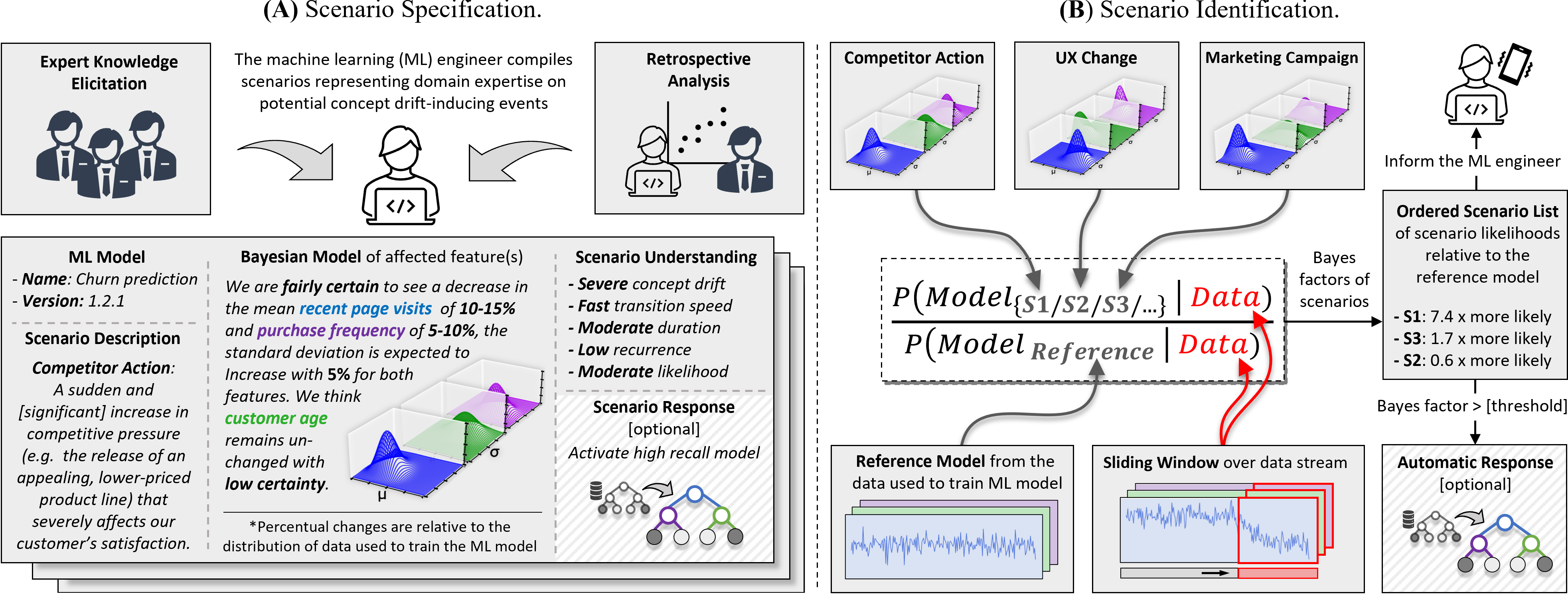}
    \caption{A visual depiction of our approach, \textit{Expert Monitoring}. In step (A), ML engineers consolidate domain expertise within the organization using a standardized format. In step (B), upon detecting feature drift, scenarios are evaluated using Bayesian model comparison. Afterwards, the ML engineer is informed about potential causes, or an automated response is triggered.}
    \label{fig:approach}
    \end{figure*}
\end{center}

\section{Approach}
We propose a method called \textit{Expert Monitoring} to address operational challenges. This method integrates domain expertise through \textit{scenario specification}. It then makes this expertise accessible to on-call ML engineers via \textit{scenario identification}, providing insights into the potential causes of feature drift upon detection (see Fig. 3).
\vspace{-0.1em}

\subsection{Scenario Specification}
% tackling model governance
We consolidate domain expertise through expert knowledge elicitation and retrospective analyses, creating a standardized resource for integration into our monitoring and response system.

\subsubsection{Expert Knowledge Elicitation}
Domain expertise is distributed among multiple experts in an organization, including marketing, product development, business strategy, and data engineering practitioners. We are interested in prior knowledge of events that can induce concept drift in the application context of an ML model. ML engineers collaborate with the domain experts to elicit scenarios of these events, using traditional requirements engineering methods such as interviews, focus groups, and observation studies~\cite{sommerville2011software}. 

\subsubsection{Retrospective Analysis}
In addition to the elicitation process, ML engineers, either in collaboration with domain experts or independently, conduct retrospective analyses. By examining the model's historical performance and correlating performance drops with feature drift events (assuming access to labeled data), they can isolate recurring problematic events for future identification.

\subsubsection{Scenario Specification Format}
The acquired scenarios are compiled by the ML engineer and stored in a standardized format. Below, we provide a description of its components.

\textbf{ML Model.} 
The name and version of the ML model that is subject to concept drift in the specified scenario.

\textbf{Scenario Description.}
This description provides the context for the event that can potentially induce concept drift.

\textbf{Bayesian Model.}
We utilize Bayesian models to provide experts an intuitive method for incorporating their prior knowledge of how the feature distribution(s) would be affected under the specified scenarios. These models are central to the next phase of our approach, namely scenario detection, detailed in Section 3.2. They enable the estimation of the distributions of the relevant feature(s) as either univariate or multivariate, simultaneously quantifying the uncertainty in the experts' subjective beliefs. Specifically, experts estimate the parameters (e.g. mean or standard deviation of an affected feature) as $\theta = Dist(location, spread)$, where the location is the estimated parameter value, and the spread indicates the expert's uncertainty. A sharp distribution implies high certainty, while a wide one suggests low certainty. For example, for a highly certain estimate that the mean customer age in the marketing campaign scenario (Fig. 2) will be eighteen, we can define a normal distribution with location 18 and spread 1.

In addition to quantifying uncertainty, the representation of domain expertise can be flexibly determined in a fine-grained manner. Experts can: (1) select alternative distributions, such as a uniform distribution, to assign equal probabilities within specific ranges (e.g., ages 16 to 20); (2) provide relative estimates, in addition to absolute ones, in relation to the ML model's training data (see Fig. 3); (3) define the distribution parameters of affected features for specific subgroups (e.g., in Fig. 1c, where the distribution of recent page visits can be estimated specifically for young people).

\textbf{Scenario Understanding.}
This includes estimating concept drift characteristics like severity, transition speed, duration, and recurrence to inform response selection (see Section 3.2). Additionally, the likelihood of the scenario, whether it is common or rare, can also be specified based on prior knowledge, using a simple three-point scale (e.g., low, moderate, high) for consistency.

\textbf{Scenario Response.}
An expert can optionally provide this response to guide the on-call ML engineer or automate scenario mitigation upon detection (see Section 3.2).

\subsection{Scenario Identification}
At runtime, upon detecting feature drift, we infer the occurrence of a scenario using the Bayesian models defined in the prior step.

\subsubsection{Bayesian Model Comparison}
Each (Bayesian) scenario model is treated as a hypothesis and has its posterior probability \( P(M|D) \) computed to assess its likelihood based on recent observations, obtained from a sliding window over the data stream. The posterior probability of a model is computed as follows:
\begin{equation}
    P(M|D) = P(M)P(D|M)
\end{equation}
Here, \( P(M) \) represents the \textit{scenario likelihood} of the model, reflecting the expert's belief about the likelihood of a scenario occurring, as discussed in Section 3.1.3. \textit{Scenario likelihoods}, which can be set to equal by default, are specified on a three-point scale and normalized to sum to one. \( P(D|M) \) denotes the \textit{marginal likelihood} and is computed using the following integral:
\begin{equation}
    P(D|M)=\int_1 \cdot\cdot\cdot \int_n P(D|\theta, M) P(\theta|M) \, d\theta_1 \cdot\cdot\cdot \, d\theta_n
\end{equation}
This represents an n-dimensional integral over all parameters $\theta$ \cite{gill2014bayesian}. However, due to the impracticality of a closed-form solution, we instead use one of the following approximation methods: (1) Markov Chain Monte Carlo sampling \cite{chib2001markov}, a computationally intensive method, or (2) calculating each feature's marginal likelihood in closed-form, by leveraging the conjugate prior assumption (the observed data and expert estimates follow the same distribution) \cite{murphy2007conjugate}, and then multiplying these likelihoods, assuming minimal inter-feature correlation.

After calculating the scenario models' posterior probabilities, we compare them with a reference model built from the observed parameter values in the ML model's training data. This comparison yields the Bayes factor \cite{gill2014bayesian}, indicating the relative likelihood of each scenario model compared to the reference model. 
%For example, a factor of 10 suggests the scenario is ten times more likely.

In assessing the role of expert estimates in Bayesian model comparison for scenario identification, we find that the precision of these estimates directly influences accuracy (Fig. 4)\footnote{For more information, see https://github.com/JoranLeest/expert\_monitoring}. Specifically, scenarios identified using low-error (small deviation from the true parameter value) and high-certainty (low standard deviation specified in the estimate) estimates are typically more accurate. Moreover, even scenarios with higher estimate errors can be correctly identified if the associated uncertainty is correctly deemed high.

\vspace{-0.65em}

\begin{center}
    \begin{figure}[!htpb]
    \centering
    \label{fig:drift_visual}
    \includegraphics[width=0.38\textwidth]{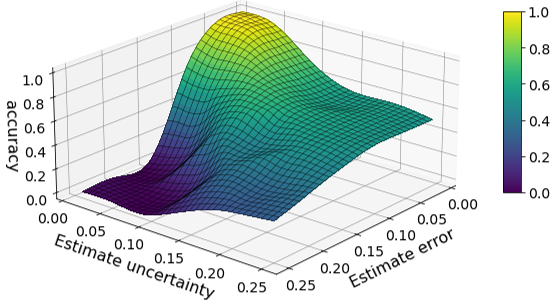}
    \caption{Detection accuracy vs. estimate uncertainty and error (in proportion relative to actual parameter values) on simulated scenarios, with a Bayes factor threshold of 5.}
    \end{figure}
\end{center}

\vspace{-2em}

\subsubsection{In Response to a Scenario}
The Bayes factors from Bayesian model comparison are displayed with scenario identifiers, assisting ML engineers in response selection for feature drift. Our method also automates concept drift mitigation by triggering the top-ranked scenario's response based on a Bayes factor threshold.

\section{Discussion}
In further developing our approach, there is an open question about extracting domain expertise on concept drift-inducing events through the reuse of requirement \cite{sommerville2011software} and knowledge \cite{shadbolt2015knowledge} elicitation methods. 
A second key question is: What is all the knowledge that we can incorporate regarding concept drift-inducing events? 
We have shown how a Bayesian model can be constructed to represent a scenario based on feature distributions, providing the base knowledge required from practitioners to explain feature drift occurrences.
However, adopting the Bayesian framework offers the flexibility allows integrating extra knowledge and extending in various directions. For instance, experts might include scenario temporal distributions and likelihood at specific times, like higher sales in then summer. 
Furthermore, our method allows for updating expert estimates and facilitates the use of scenario-specific machine learning models, both of which help mitigate recurring scenarios.

% Focus on external events, data integirty issues less predictable, especially valuable in critical domains and for core models, where performance deteroiration needs to be quickyl resolved. 

% Scenario identification and response selection can be set as fuzzy rules, for example, competitor actions can have a varying impact on customer satisfaction, despite affecting the same features, the features will change to varying degrees. This way we can also choose milder responses for mild competitor actions and prompt responses for more severe competitor actions.

The literature reveals a gap in understanding human-in-the-loop systems, especially domain expertise, for concept drift challenges. This area warrants more study to aid practitioners with appropriate processes, tools, and methodologies. While previous works offer methods for detecting feature drift-related model failures in practical settings \cite{rabanser2019failing, sethi2017reliable, shankar2021towards, garg2022leveraging, guillory2021predicting}, they overlook the latent issue of concept drift. More closely related to our research, Chen et al. \cite{chen2021mandoline} and Cobb et al. \cite{cobb2022context} incorporate domain expertise in performance estimation and feature drift detection, but also do not address concept drift. Our work uniquely integrates domain expertise in identifying and addressing concept drift, contributing to human-centered model monitoring \cite{shergadwala2022human}. We believe that scenario-based methods like ours, similar to those used in software architecture \cite{leest2023evolvability}, are promising for advancing model monitoring.

%There is a noticeable gap in the literature of how a human-in-the-loop, and in particular domain expertise, can actively contribute in addressing concept drift. We believe this area of research warrants more attention to offer ML engineers comprehensive support in the form of processes, methodologies, and tooling. Prior research on developing automated methods for concept drift detection without labeled data \cite{sethi2017reliable, dos2016fast, gozuaccik2019unsupervised, gulcan2023unsupervised} lacks practical context integration, resulting in limitations tied to the opacity of model performance and latent variables wherein concept drift emerges. Some methods for concept drift detection allow for [implicit] incorporation of domain expertise in the selection of hyperparameters, like the size of the sliding window size by considering the expected abruptness \cite{gama2014survey} or using knowledge of drift severity and duration in change point detection \cite{bach2010bayesian}. Our work is the first to [explicitly] integrate domain expertise in a method for detecting and mitigating concept drift, making an initial contribution to the emerging field of human-centered model monitoring \cite{shergadwala2022human}. We believe scenario-based methods like ours represent a promising avenue for advancing model monitoring, similar to their prior application in software architecture for making decisions about evolvability concerns \cite{leest2023evolvability}.

\section{Future plans}
We contend that the inherent intricacy of the human factors involved in our approach needs to be addressed through rigorous evaluation and collaboration with practitioners. Specifically, we identified the following research questions:
(1) What domain expertise of concept drift-inducing events can be elicited, and represented with sufficient detail? (2) Can scenarios be identified through Bayesian model comparison with sufficient accuracy? (3) Is the \textit{Expert Monitoring} approach perceived as helpful by ML engineers and does it enable them to improve on business-related metrics?
To answer (1) and (2), we'll conduct action research via workshops and focus groups with diverse domain practitioners, refining our approach based on real-world industrial needs. For (3), we'll use surveys and interviews to gauge our approach's usefulness.

\section*{Acknowledgment}
This research is supported by ExtremeXP, a project co-funded by the European Union Horizon Programme under Grant Agreement No. 101093164.

\bibliographystyle{ACM-Reference-Format}
\bibliography{bibliography.bib}

\end{document}